\documentclass{article}
\usepackage{graphicx} 
\usepackage{amsfonts,amsmath}
\usepackage{caption}
\usepackage{subcaption}
\usepackage{verbatim}
\usepackage[ruled,vlined]{algorithm2e}

     \usepackage[preprint,nonatbib]{neurips_2024}

\usepackage[utf8]{inputenc} 
\usepackage[T1]{fontenc}    
\usepackage{hyperref}       
\usepackage{url}            
\usepackage{booktabs}       
\usepackage{amsfonts}       
\usepackage{nicefrac}       
\usepackage{microtype}      
\usepackage{xcolor}         
\title{Noisy Data Visualization using Functional Data Analysis}

\begin{document}

\global\long\def\x{\mathbf{x}}
\global\long\def\z{\mathbf{z}}
\global\long\def\u{\mathbf{u}}
\global\long\def\v{\mathbf{v}}
\global\long\def\m{\mathbf{m}}
\global\long\def\y{\mathbf{y}}
\global\long\def\h{\mathbf{h}}
\global\long\def\s{\mathbf{s}}
\global\long\def\E{\mathbb{E}}
\

\author{%
  Haozhe~Chen \\
  Department of Mathematics \& Statistics\\
  Utah State University\\
  Logan, UT, USA\\
  \texttt{a02314155@usu.edu} \\
  \And
 Andres Felipe Duque Correa \\
  Department of Mathematics \& Statistics\\
  Utah State University\\
  Logan, UT, USA \\
  \texttt{andres.duque@usu.edu}\\
  \AND
  Guy~Wolf \\
  Department of Mathematics and Statistics \\
  University of Montreal \\
  Montreal, QC, Canada\\
  \texttt{wolfguy@mila.quebec} \\
  \And
  Kevin R.~Moon \\
  Department of Mathematics \& Statistics\\
  Utah State University\\
  Logan, UT, USA \\
  \texttt{kevin.moon@usu.edu} \\
}

\maketitle

\begin{abstract}
Data visualization via dimensionality reduction is an important tool in exploratory data analysis. However, when the data are noisy, many existing methods fail to capture the underlying structure of the data. The method called Empirical Intrinsic Geometry (EIG) was previously proposed for performing dimensionality reduction on high dimensional dynamical processes while theoretically eliminating all noise. However, implementing EIG in practice requires the construction of high-dimensional histograms, which suffer from the curse of dimensionality. Here we propose a new data visualization method called Functional Information Geometry (FIG) for dynamical processes that adapts the EIG framework while using approaches from functional data analysis to mitigate the curse of dimensionality. We experimentally demonstrate that the resulting method outperforms a variant of EIG designed for visualization in terms of capturing the true structure, hyperparameter robustness, and computational speed.  We then use our method to visualize EEG brain measurements of sleep activity. 
    
\end{abstract}

\section{Introduction} 
\label{sec:Introduction} 
High-dimensional datasets often contain redundant information, resulting in an artificially elevated extrinsic dimension. However, their underlying structure can often be accurately modeled as a low-dimensional manifold with some added noise. In machine learning, the concept of a manifold loosely refers to a connected set of points in high-dimensional space that can be effectively approximated using a lower number of dimensions. Building upon the manifold assumption, manifold learning methods aim to learn a lower-dimensional representation of data while retaining as much of its inherent information as possible. Manifold learning has been useful in many fields such as image classification and object detection~\cite{lezcano2019trivializations, verma2019manifold, rodriguez2020embedding}, image synthesis and enhancement~\cite{dai2022adaptive, luo2022progressive}, video analysis~\cite{zhen2019dilated, wang2022dreamnet}, 3D data processing~\cite{porikli2010learning}, analyzing single-cell RNA-sequencing data~\cite{moon2018manifold}, and more~\cite{fassold2023survey}

In particular, manifold learning has been used in nonlinear dimensionality reduction and data visualization. Classical approaches for nonlinear dimension reduction are Isomap ~\cite{balasubramanian2002isomap}, MDS~\cite{cox2000multidimensional}, Local Linear Embedding (LLE)~\cite{roweis2000nonlinear} and Laplacian Eigenmaps~\cite{belkin2003laplacian}. In recent years, more powerful approaches like t-SNE~\cite{van2008visualizing}, UMAP~\cite{mcinnes2018umap} have emerged. However, in many applications, the noise level may make it difficult to learn the true, low-dimensional structure of the data. Many of these popular manifold learning and data visualization methods fail to take noise directly into account. A more advanced method called diffusion maps (DM)~\cite{coifman2006dm} has been proposed that denoises and learns the manifold structure by using a Markov diffusion process to emphasize data connectivity and preserve the intrinsic geometry. However, DM is not optimal for visualization because it tends to encode information in more than 2 or 3 dimensions~\cite{haghverdi2016diffusion,phate}. More recently, PHATE\cite{phate} has been used to visualize noisy data. Like DM, PHATE also uses a diffusion framework to learn the global structure and denoise, but is better designed for data visualization. 

Despite the denoising capabilities of DM and PHATE, the noise levels in some types of data may still be difficult to overcome completely. However, if the data can be assumed to be measured from a dynamical system, additional structure can be leveraged.
Dynamical systems and time series data can often be viewed as processes governed by a small set of underlying parameters, effectively confined within a low-dimensional manifold. Under certain assumptions on the generating process, a distance metric can be learned between points using the Empirical Intrinsic Geometry (EIG) framework that is theoretically noise-free under various forms of noise~\cite{eig,eig1}. To do this, EIG leverages the fact that the Mahalanobis distance is invariant to linear transformations. The authors  showed that various forms of noise (e.g. additive and multiplicative) result in linear transformations in the probability space. Computing the Mahalanobis distance between empirically constructed local histograms thus results in a distance that is close to the true distance between the underlying parameters that drive the system~\cite{eig,eig1}. Dynamical Information Geometry (DIG)~\cite{dig} was then later introduced to visualize the data from these distances using a diffusion and information distance framework~\cite{phate,coifman2006dm}.

Despite their nice theoretical properties, for finite samples EIG and DIG suffer limitations due to their use of local histograms to transform to the probability space.  In particular, histograms are difficult to construct on data with more than two dimensions as the number of bins in the histogram increases exponentially with dimension. This results frequently in sparsely populated histograms, which also makes local covariance matrix estimation difficult. 

To overcome these difficulties, we revisit the EIG framework by considering the local probability densities directly. Thus our contributions are the following: 1) Using a functional data analysis framework~\cite{galeano2015mahalanobis, ramsay2005principal}, we show that pairwise Mahalanobis distances can be constructed between probability distributions without directly estimating the densities, thus mitigating the curse of dimensionality that afflicts the histogram approach. To do this, some notion of neighbors must be given that indicates which points are neighbors of each other. We focus primarily on time series data in which neighbors are determined by a time window, although other neighborhood definitions could be used.
2) We then embed these distances into low dimensions using a diffusion and information geometry framework, which is useful for data visualization. We call this visualization method Functional Information Geometry (FIG). 3) We  demonstrate FIG on simulated time series data and real data taken from EEG measurements of the brain during different sleep stages. We show that when compared to DIG and EIG, FIG better captures the true low-dimensional structure, is more robust to hyperparameters, and is computationally faster.



\section{Problem Setting and Background}
\label{sec:background}
 We use the same state-space formalism given in EIG~\cite{eig,eig1} and DIG~\cite{dig} for time series data:
\begin{align}
    \x_t &= \y_t (\boldsymbol \theta_t) + \boldsymbol \xi_t \label{eq:dynamicalsystem1} \\
    d\theta_t^i &= a^i(\theta_t^i) + d w_t^i, i = 1, ..., d. \label{eq:dynamicalsystem2}
\end{align}
$\x_t$ represents the observed multivariate time series that is a noisy version of $\y_t$. The parameters $\boldsymbol 
\theta_t$ represent the hidden states that
drive the clean process $\y_t$. The noise $\boldsymbol \xi_t$ is a stationary process independent
of $\y_t$. Each $\y_t$ is drawn from a conditional pdf $p(\boldsymbol y| \boldsymbol \theta)$. We assume that the parameters are affected by the unknown drift functions $a^i$ that are independent from $\theta_j$ when $j \neq i$. This allows us to assume local independence between $\theta_t^i$ and $\theta_t^j$, $\forall j \neq i$. Finally, the $w_t^i$ variables are governed by a Brownian motion process. 

The goal is to derive a distance that approximates pairwise distances between the parameters $\boldsymbol 
\theta_t$ that drive the process. The density $p(\x|\boldsymbol \theta)$ is a linear transformation of the density $p(\y|\boldsymbol \theta)$~\cite{eig,eig1}.  Since the densities are unknown, the authors in~\cite{eig,eig1,dig} use histograms as their estimators. Each histogram $\h_t = (h_t^1, ..., h_t^{N_b})$ has $N_b$ bins and is built with the observations within a time window of length $L_1$, centered at $\x_t$. It was also shown that the expected value of the histograms $\mathbb E(h_t^j)$, is a linear transformation of $p(\x|\boldsymbol \theta)$~\cite{eig,eig1}. Since the Mahalanobis distance is invariant under linear transformations, it can be deduced that the following distance is noise resilient: 
\begin{align}
    d^2 (\x_t, \x_s) = \big( \mathbb{E} [\h_t] - \mathbb{E} [\h_s] \big)^T\big(\widehat{C}_t + \widehat{C}_s \big)^{-1} \big(\mathbb{E} [\h_t] - \mathbb{E} [\h_s] \big) \label{eq:histdist}
\end{align}
where $\widehat{C}_t$ and $\widehat{C}_s$ are the local covariance matrices of the histograms. They are constructed using a time window centered at $\boldsymbol h_t$ and $\boldsymbol h_s$, respectively. Thus under certain assumptions, $d^2(\x_t, \x_s)$ is a good approximation of the distance between the underlying state variables~\cite{eig1}, eliminating the effects of noise; i.e.  $ \|\boldsymbol \theta_t - \boldsymbol \theta_s\|^2 \simeq d^2(\x_t, \x_s)$. Our primary goal is to derive an alternative Mahalanobis distance that still approximates the distances between the parameters $\boldsymbol \theta_t$ but avoids constructing histograms or any other density estimator.

Our newly derived distance can be used as the input to various machine learning algorithms. As we are focused on data visualization, we choose to input the distances into PHATE~\cite{phate}. PHATE first converts pairwise distances into local affinities  by employing an $\alpha$-decay kernel with adaptive bandwidth, allowing affinities to adjust based on local density, while correcting inaccuracies in sparsely sampled regions. Diffusion is then performed on the corresponding affinity graph, which learns the global relationships between points while denoising. PHATE then extracts the information learned in the diffusion process for visualization by constructing potential distances from diffused probabilities and directly embedding them into low dimensions using metric MDS~\cite{cox2000multidimensional}, avoiding instability and inaccuracies often encountered with direct embedding of diffusion distances. The resulting visualization tends to better represent the local and global structure of noisy data than competing methods~\cite{phate}. Several adaptations have PHATE have been created for visualizing supervised data problems~\cite{rhodes2021random}, visualizing data at multiple scales~\cite{kuchroo2022multiscale}, and visualizing the internal geometry of neural networks~\cite{gigante2019visualizing}.

\section{Functional Information Geometry}
\label{sec:FIG}
Constructing the histograms and covariance matrices in the distance in~\ref{eq:histdist} for high dimensional data is difficult due to the curse of dimensionality. Since the noise causes a linear transformation in the probability space, we can obtain another distance that is noise resilient that avoids estimating the probability density by using functional data analysis. 
For our approach, we will need to define the Mahalanobis distance between functions. To do this, we will use concepts from~\cite{galeano2015mahalanobis,ramsay2005principal}. However, we cannot directly use the functional data analysis (FDA) framework from~\cite{galeano2015mahalanobis,ramsay2005principal} for our problem. Instead, multiple modifications are required for the following reasons: 1) We are considering probability densities centered at the data points as the functions whereas standard FDA attempts to learn a function directly from the data. 2) The densities may have a multivariate input. 3) We need to define the Mahalanobis distance in the case where the two points (functions or densities) come from different distributions.

\subsection{Vector Mahalanobis Distance}
We will first define the vector Mahalanobis distance in terms of principal components as the functional Mahalanobis distance will similarly use functional principal components. Suppose we have two points $\u,\v\in\mathbb{R}^d$ that have the same distribution with the same covariance matrix $C$ and mean vector $\m$. Then the Mahalanobis distance between $\u$ and $\v$ is 
\[
d_M(\u,\v)=[(\u-\v)^TC^{-1}(\u-\v)]^{1/2}.
\]
 This is the distance obtained after standardizing the data. Since the two points share the same mean, the mean does not play a role in this specific distance. Consider the eigendecomposition $C=V\Lambda V^T$. The vectors of principal component scores are given by $\s_u=V^T (\u-\m)$ and $\s_v=V^T(\v-\m)$. Thus $\u=\m+V\s_u$ and $\v=\m+V\s_v$. Plugging this into the Mahalanobis distance gives 
\begin{align}
    d_M(\u,\v)&=[(\s_u-\s_v)^T V^T V \Lambda^{-1}V^T V(\s_u-\s_v)]^{1/2} \nonumber \\
    &=||\Lambda^{-1/2}(\s_u-\s_v)||. \label{eq:M_pca}
\end{align}
This method for computing the Mahalanobis distance was given in~\cite{ramsay2005principal}. We will now extend this to the case where $\u$ and $\v$ have different distributions. Let the covariance matrices and means be $C_u,$  $C_v$, $\m_u,$ and $\m_v$, respectively. In \cite{eig, dig}, the joint covariance between two observations was defined as $(C_u+C_v)$. Thus the Mahalanobis distance is given by:
\begin{equation}
d_M(\u,\v)=[(\u-\v)^T(C_u+C_v)^{-1}(\u-\v)]^{1/2}. \label{eq:def}
\end{equation}
We will need an expression of $(C_u+C_v)^{-1}$ in terms of principal components. In \cite{eig_pre}, a Taylor expansion around the observable variables $\u$ and $\v$ are given, which yields the second-order approximation of the Euclidean distance between unobservable hidden processes $\boldsymbol \theta_u$ and $\boldsymbol \theta_v$:
\begin{align}
    \|\boldsymbol \theta_u - \boldsymbol \theta_v\|^2 = \frac{1}{2}(\u - \v)^T (C_u^{-1} + C_v^{-1})(\u - \v) + \mathcal{O}(\|\u-\v\|^4).
\end{align}
We can thus instead define the vector Mahalanobis distance as: 
\begin{equation}
d_M(\u,\v)= [(\u-\v)^T(C_u^{-1}+C_v^{-1})(\u-\v)]^{1/2}. \label{eq:def1}
\end{equation}
As this is similar to the sum of two standard Mahalanobis distances, the principal components version simply has two of the terms in Eq.~\ref{eq:M_pca}. In this case, let $C_u=V_u\Lambda_uV_u^T$ and $C_v=V_v\Lambda_vV_v^T$. The principal component scores are given by $\s_{uu}=V_u^T (\u-\m_u)$, $\s_{vv}=V_v^T(\v-\m_v)$, $\s_{uv}=V_u^T (\v-\m_u)$, $\s_{vu}=V_v^T(\u-\m_v)$. Then the Mahalanobis distance is:
\begin{equation}
    d_M(\u,\v)=\left(||\Lambda_u^{-1/2}(\s_{uu}-\s_{uv})||^2+ ||\Lambda_v^{-1/2}(\s_{vu}-\s_{vv})||^2\right)^{1/2}.
\end{equation}

\subsection{Functional Mahalanobis Distance Between Densities}
We now look to extend the previous distances to the function setting where the functions are probability densities. Let $f$ be a functional random variable that is also a probability density and is in $L^2(T)$, where $T\subseteq \mathbb{R}^d$. 
Let $\mu_f(t)=\E[f(t)]$ be the density mean and a covariance operator $\Gamma_f$ be 
\begin{equation}
    \Gamma_f(\eta)=\E[(f-\mu_f)\otimes(f-\mu_f)(\eta)],
\end{equation}
where for any $\eta\in L^2(T)$,
\begin{equation}
    (f-\mu_f)\otimes(f-\mu_f)(\eta)=\langle f-\mu_f,\eta \rangle (f-\mu_f),
\end{equation}
 where we use the $L^2$ inner product:
 \[
 \langle f-\mu_f,\eta \rangle =\int_T (f(x)-\mu_f(x))\eta(x)dx.
 \]
 In \cite{galeano2015mahalanobis}, when $\Gamma_f$ exists, then there exists a sequence of non-negative eigenvalues of $\Gamma_f$, denoted as $\lambda_1\geq \lambda_2 \geq \cdots,$ with $\sum_{k=1}^\infty \lambda_k <\infty$, and a set of orthonormal eigenfunctions denoted as $\psi_1,\psi_2,\dots$ such that $\Gamma_f(\psi_k)=\lambda_k\psi_k$ for all $k$. The eigenfunctions form an orthonormal basis of $L^2(T)$ so we can write $f$ in terms of this basis as:
 \begin{equation}
     f=\mu_f+\sum_{k=1}^\infty \theta_k\psi_k,
 \end{equation}
where $s_k=\langle f-\mu_f,\psi_k\rangle$ are the functional principal component scores of $f$. Since the eigenfunctions are orthonormal, then the functional principal component scores are uncorrelated with zero mean and variance $\lambda_k$.

We will need the inverse covariance operator $\Gamma_f^{-1}$. From \cite{galeano2015mahalanobis,ramsay2005principal}, a regularized version of it is defined as:
\begin{equation}
    \Gamma_K^{-1}(\xi)=\sum_{k=1}^K \frac{1}{\lambda_k}(\psi_k\otimes \psi_k)(\xi),
\end{equation}
where $K$ is chosen as some threshold and $\xi$ is in the range of $\Gamma$. Then based on Eq. (\ref{eq:M_pca}), if two densities $f_i$ and $f_j$ come from the same distribution (i.e. have the same mean function and covariance operator), then the principal component version of the Mahalanobis distance is
\begin{equation}
d_{FM}^2(f_i,f_j)=\sum_{k=1}^K(\omega_{ik}-\omega_{jk})^2,
\label{eq:Mahalanobis distance}
\end{equation}
where $\omega_{ik}=s_{ik}/\lambda_k^{1/2}$~\cite{ramsay2005principal}.

We now extend to the case when $f_i$ and $f_j$ have different distributions. Let $\Gamma_i$ and $\Gamma_j$ be the two covariance operators. Their inverses are defined as before with eigenvalues $\lambda_{ik}$ and $\lambda_{jk}$. Define the following functional principal component scores:
\begin{align}
    s_{iik}&=\langle f_i-\mu_i,\psi_{ik}\rangle \nonumber \\ 
    s_{ijk}&=\langle f_j-\mu_i,\psi_{ik}\rangle  \nonumber \\
    s_{jik}&=\langle f_i-\mu_j,\psi_{jk}\rangle \nonumber \\
    s_{jjk}&=\langle f_j-\mu_j,\psi_{jk}\rangle.  
\end{align}
Then the Mahalanobis distance becomes
\begin{equation}
     d_{FM}^2(f_i,f_j)= \sum_{k=1}^K(\omega_{iik}-\omega_{ijk})^2 + \sum_{k=1}^K(\omega_{jik}-\omega_{jjk})^2,
     \label{eq:finaldist}
\end{equation}
where $\omega_{ijk}=s_{ijk}/\lambda_{ik}^{1/2}$.

\subsection{Learning the Principal Component Scores}
All that remains is to derive the functional principal component scores $s_{ijk}$. Standard FDA methods, such as Functional Principal Components Analysis (FPCA), typically require basis functions (e.g. the Fourier basis or a spline basis) to be fit to the data. While we do employ basis functions in the following, we are able to exploit properties of probability densities by computing empirical averages of the basis functions, avoiding the need for directly estimating the densities. Thus our approach differs from that of standard FPCA.

Suppose we have densities $f_1,\dots,f_n$ and some notion of ``neighbors" of these densities. In practice, we model each measured data point $\x_i$ as being drawn from $f_i$ and the neighbors of $f_i$ are determined by the corresponding data points that are within a time window with fixed length $L$ with $\x_i$ at the center. However, we note that neighbors could be determined in other ways such as the Euclidean distance between the data points. This means that our proposed distance can be computed as long as some notion of neighbors between data points is defined.

Assume for now that the densities $f_1,\dots,f_n$ are known. We will relax this assumption later. Then the mean function of the density $f_i$ can be estimated as 
\begin{equation}
\mu_{f_i}(\x)=\frac{1}{|\mathcal{N}_i|}\sum_{j\in \mathcal{N}_i} f_j(\x),
\end{equation}
where $\mathcal{N}_i$ denotes the set of indices such that $f_j$ is a neighbor of $f_i$. In real-world applications, we have the flexibility to employ distinct window sizes for various purposes, such as computing the average of basis functions or calculating covariance matrices, as exemplified in detail in Algorithm~\ref{FigAlgorithm}. To simplify the presentation, we use the same set of neighbors (and therefore window size) for all of these tasks.
An estimate of the the covariance function associated with $f_i$ is then
\begin{equation}
    v_i(\x,\z)=\frac{1}{|\mathcal{N}_i|}\sum_{j\in \mathcal{N}_i} (f_j(\x)f_j(\z)-\mu_{f_i}(\x)\mu_{f_i}(\z)).
\end{equation}
We will need to find the eigenfunction of this which satisfies 
\begin{equation}
    \int v_i(\x,\z)\psi_i(\z)d\z=\lambda_i \psi_i(\x).
    \label{eq:eigenfunction}
\end{equation}
We will assume a basis expansion for the eigenfunction $\psi_i(\x)=\phi(\x)^T\mathbf{b}_i$ where $\phi:\mathbb{R}^d\rightarrow\mathbb{R}^M$ is a set of basis functions such as the Fourier basis or cubic splines. The eigenfunction equation becomes
\begin{align}
    \lambda_i \phi(\x)^T\mathbf{b}_i &=\int v_i(\x,\z)\psi(\z)d\z \nonumber \\
    &=\frac{1}{|\mathcal{N}_i|}\sum_{j\in \mathcal{N}_i}\int (f_j(\x)f_j(\z)-{\mu}_{f_i}(\x){\mu}_{f_i}(\z))\phi(\z)^T\mathbf{b}_id\z \nonumber \\ 
    &= \left[ \frac{1}{|\mathcal{N}_i|}\sum_{j\in \mathcal{N}_i} f_j(\x) \int f_j(\z)\phi(\z)^Td\z - \mu_{f_i}(\x)\frac{1}{n}\frac{1}{|\mathcal{N}_i|}\sum_{j\in \mathcal{N}_i} \int f_j(\z) \phi(\z)^Td\z\right]\mathbf{b}_i \nonumber \\
    &= \left[ \frac{1}{|\mathcal{N}_i|}\sum_{j\in \mathcal{N}_i} f_j(\x)\mathbf{a}_j^T- \mu_{f_i}(\x)\frac{1}{|\mathcal{N}_i|}\sum_{j\in \mathcal{N}_i} \mathbf{a}_j^T\right]\mathbf{b}_i, \label{eq:eigenfunction2}
\end{align}
where $\mathbf{a}_i=\int f_i(\z) \phi(\z)d\z$. The vector $\mathbf{a}_i$ is the expected value of the basis function $\phi$ with respect to the density $f_i$. This value is easily approximated using a sample mean as 
\begin{align}
    \hat{\mathbf{a}}_i=\frac{1}{|\mathcal{N}_i|}\sum_{j\in \mathcal{N}_i} \phi(\x_j). 
    \label{eq:compute_ai}
\end{align}
Define $W=\int\phi(\x)\phi(\x)^Td\x$. If $\phi$ is an orthonormal basis (such as the Fourier basis), then $W$ is the identity matrix. Now multiply both sides of Eq.~\ref{eq:eigenfunction2} by $\phi(\x)$ and then integrate to obtain:
\begin{align}
    \lambda_i \int \phi(\x)\phi(\x)^Td\x \mathbf{b}_i&= \lambda_i W\mathbf{b}_i \nonumber \\
    &= \int\phi(\x)\left[ \frac{1}{|\mathcal{N}_i|}\sum_{j\in \mathcal{N}_i} f_j(\x)\mathbf{a}_j^T- \mu_{f_i}(\x)\frac{1}{|\mathcal{N}_i|}\sum_{j\in \mathcal{N}_i} \mathbf{a}_j^T\right]\mathbf{b}_i d\x \nonumber \\
    &= \left[ \frac{1}{|\mathcal{N}_i|}\sum_{j\in \mathcal{N}_i} \mathbf{a}_j\mathbf{a}_j^T- \frac{1}{|\mathcal{N}_i|}\sum_{j\in \mathcal{N}_i} \mathbf{a}_j\frac{1}{|\mathcal{N}_i|}\sum_{j\in \mathcal{N}_i} \mathbf{a}_j^T\right]\mathbf{b}_i \nonumber \\
    &= \left[ \frac{1}{|\mathcal{N}_i|}\sum_{j\in \mathcal{N}_i} \mathbf{a}_j\mathbf{a}_j^T- \mu_{a_i}\mu_{a_i}^T\right]\mathbf{b}_i \nonumber \\
    &=\tilde{A}_i\mathbf{b}_i
    \label{eq:compute_A}
\end{align}
where $\mu_{a_i}=\frac{1}{|\mathcal{N}_i|}\sum_{j\in \mathcal{N}_i} \mathbf{a}_j$ and $\tilde{A}_i=\frac{1}{|\mathcal{N}_i|}\sum_{j\in \mathcal{N}_i} \mathbf{a}_j\mathbf{a}_j^T- \mu_a\mu_a^T$. The matrix $\tilde{A}_i$ can be viewed as the sample covariance matrix of the vector $\mathbf{a}_i$. By  letting $\mathbf{u_i}=W^{1/2}\mathbf{b}_i$, we get the following eigen equation:
\begin{equation}
    W^{-1/2}\tilde{A}_iW^{-1/2}\mathbf{u}_i=\lambda_i\mathbf{u}_i.\label{eq:eigenequation}
\end{equation}
Now let $\hat{\mu}_i=\frac{1}{|\mathcal{N}_i|}\sum_{j\in \mathcal{N}_i} \hat{\mathbf{a}}_j$ and approximate $\tilde{A}_i\approx\frac{1}{|\mathcal{N}_i|}\sum_{j\in\mathcal{N}_i} \hat{\mathbf{a}}_j\hat{\mathbf{a}}_j^T-\hat{\mu}_j\hat{\mu}_j^T$. If we find the first $K$ eigenvalues and eigenvectors of Eq.~\ref{eq:eigenequation}, which we denote individually as $\lambda_{ik}$ and $\mathbf{u}_{ik}$ for $k=1,\dots,K$, we then obtain the projections  
\begin{equation}
    s_{ijk}=(\hat{\mathbf{a}}_j-\hat{\mu}_i)^T W^{-1/2}\mathbf{u}_{ik},
    \label{eq:compute_s}
\end{equation}
We then get $\omega_{ijk}=s_{ijk}/\lambda_{ik}^{1/2}$, which is used in Eq.~\ref{eq:finaldist} to obtain the final distance. This distance will then accurately approximate  the distance between the parameters $\boldsymbol{\theta_t}$ as long as the approximations of $\mathbf{a}_i$ and $\tilde{A}_i$ are accurate. To embed the final distances into low dimensions, especially for visualization, we input the distances into the PHATE algorithm to obtain the final FIG embedding.

 We reiterate that an advantage of FIG is that we do not need to estimate the densities (in contrast with EIG and DIG) nor do we need to fit the densities to a basis. We simply need to compute the $\mathbf{a}_i$ vectors which are obtained by taking an empirical average of the basis functions. This becomes easy to extend to higher dimensions as well. One final consideration is the numerical stability of the eigenvalues $\lambda_{ik}$. In our experiments, we frequently found that some eigenvectors $\mathbf{u}_{ik}$ are structurally informative but have relatively low eigenvalues. Dividing by the square root of these eigenvalues tends to amplify numerical errors. In contrast, exponentiation is less prone to such error amplification while still providing a similar function shape as $\lambda^{-1/2}$, making it a more stable choice. Hence we use the following normalized principal component scores: 
 \begin{equation}
     \omega_{ijk}=s_{ijk}/ e^{\lambda_{ik}}.
     \label{eq:eigenvec_normalization}
 \end{equation}
See Alg.~\ref{FigAlgorithm} for a summary of all of the steps in FIG when applying it to time series data.

\begin{algorithm}
\SetAlgoLined
\SetKwInput{KwData}{Input}
\KwData{Input data $X = \{\x_1,\dots,\x_n\}$ ordered by time, basis function $\phi$, time window centered at time $\x_t$ with length $\mathcal{L}_1$, window size $\mathcal{L}_2$ to compute the covariance, desired embedding dimension $r$ (usually 2 or 3 for visualization)}
\KwResult{The FIG embedding $Z_r$}
\textbf{1:} $\phi(\x_i) \leftarrow$ compute the  basis (e.g. Fourier) for each sample $\x_i$ 

\textbf{2:} $\hat{\mathbf{a}}_i \leftarrow$ compute the new feature representations using Eq. \ref{eq:compute_ai} using a window of points with size  $\mathcal{L}_1$ centered at $\x_i$. 

\textbf{3:} $\hat{A}_i \leftarrow$ compute the covariance matrix as $\tilde{A}_i=\frac{1}{\mathcal{L}_2}\sum_{j\in\mathcal{W}_i} \hat{\mathbf{a}}_j\hat{\mathbf{a}}_j^T-\hat{\mu}_j\hat{\mu}_j^T$, and $\hat{\mu}_j=\frac{1}{\mathcal{L}_2}\sum_{k\in \mathcal{W}_j} \hat{\mathbf{a}}_k$, where $\mathcal{W}_i$, $\mathcal{W}_j$ are the windows of points centered at $\x_i$ and $\x_j$  respectively with size $\mathcal{L}_2$.

\textbf{4:} $s_{ijk} \leftarrow$ apply the eigen-decomposition to $\tilde{A}_i$, using Eq.~\ref{eq:eigenequation} and Eq.~\ref{eq:compute_s} to obtain the obtain the principal components.  

\textbf{5:} $w_{ijk} \leftarrow$ normalize the principal components by Eq.~\ref{eq:eigenvec_normalization} 

\textbf{6:} $\mathcal{D}_{FM} \leftarrow$ compute the functional Mahalanobis distance matrix in Eq.~\ref{eq:finaldist} from $w_{ijk}$

\textbf{7:} Input $\mathcal{D}_{FM}$ into the PHATE algorithm to obtain the embedding $Z_r$
\caption{The FIG Algorithm for time series data}
\label{FigAlgorithm}
\end{algorithm}

\section{Experiments}
\label{sec:experiments}
\subsection{Simulated study}
\begin{figure}
\centering
  \begin{subfigure}{.5\textwidth}
  \centering
  \includegraphics[width=.8\linewidth]{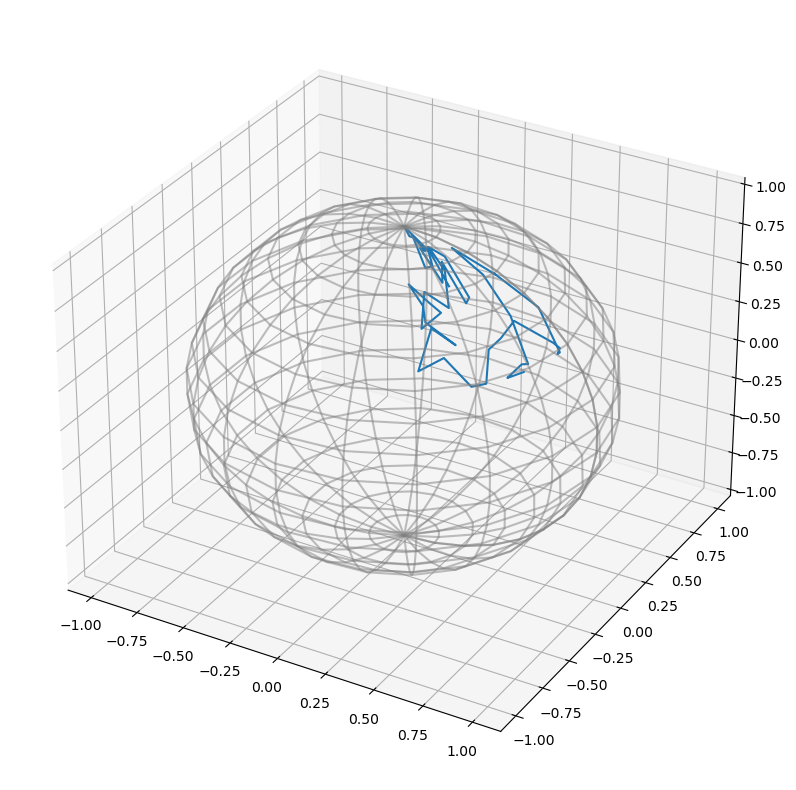}
\end{subfigure}%
\begin{subfigure}{.5\textwidth}
  \centering
  \includegraphics[width=.9\linewidth]{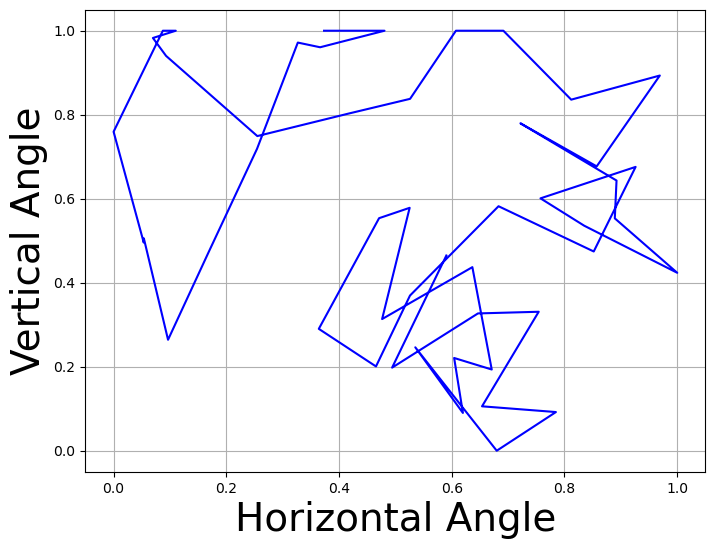}
\end{subfigure}
  \caption{Simulated data setup. (Left) Segment of the 3D movement of the object on the sphere. (Right) Corresponding segment of the 2D trajectory of the two independent factors: the horizontal and vertical angles.}
  \label{fig:randomwalk}
\end{figure}

For this study, inspired by the work of \cite{eig}, we conducted simulations to mimic the movement of a radiating object across a 3D sphere. The motion is primarily governed by two angles: the horizontal (azimuth) angle $\theta_t^1$ and the vertical (elevation) angle $\theta_t^2$. Building on the assumptions outlined in \cite{eig}, we demonstrate that the observed movements in 3D space can be understood through the vector $\boldsymbol{\theta} = [\theta^1_t, \theta_t^2]$. Figure \ref{fig:randomwalk} showcases a segment of our simulation. Referring to Equations \ref{eq:dynamicalsystem1} and \ref{eq:dynamicalsystem2}, the left side of the figure displays the clean measurements $\y_t$ in 3D space, while the right side represents the 2D underlying process, $\boldsymbol{\theta}$. In our experiments, we simulated 1000 random steps, adding noise $\boldsymbol{\xi}_t$ as described in Equation \ref{eq:dynamicalsystem1}. We use Algorithm~\ref{FigAlgorithm} with 7 Fourier basis functions for each dimension of the data to derive the distance matrix and obtain a 2D FIG embedding.
. 

We add random Gaussian noise with a mean of zero, and the noise level is determined solely by the standard deviation $\sigma$. To assess noise robustness, we compared FIG with other baseline visualization methods: DIG, PHATE, UMAP, and t-SNE, across different noise levels. We assume the noise is independent and identically distributed (i.i.d.). We systematically increased the noise until the Mantel coefficient (described below) between the noisy data and $\boldsymbol{\theta}$ dropped below $0.5$. All embeddings are obtained as 2-dimensions. For PHATE, UMAP, and t-SNE, we use the Python APIs and libraries: \href{https://phate.readthedocs.io/en/stable/}{phate}, \href{https://umap-learn.readthedocs.io/en/latest/}{umap}, \href{https://scikit-learn.org/stable/modules/generated/sklearn.manifold.TSNE.html}{t-sne}. For DIG and FIG, we take $\mathcal{L}_1 = 10$, and $\mathcal{L}_2 = 10$. 

\begin{figure}
\centering
  \centering
  \includegraphics[width=.73\textwidth]{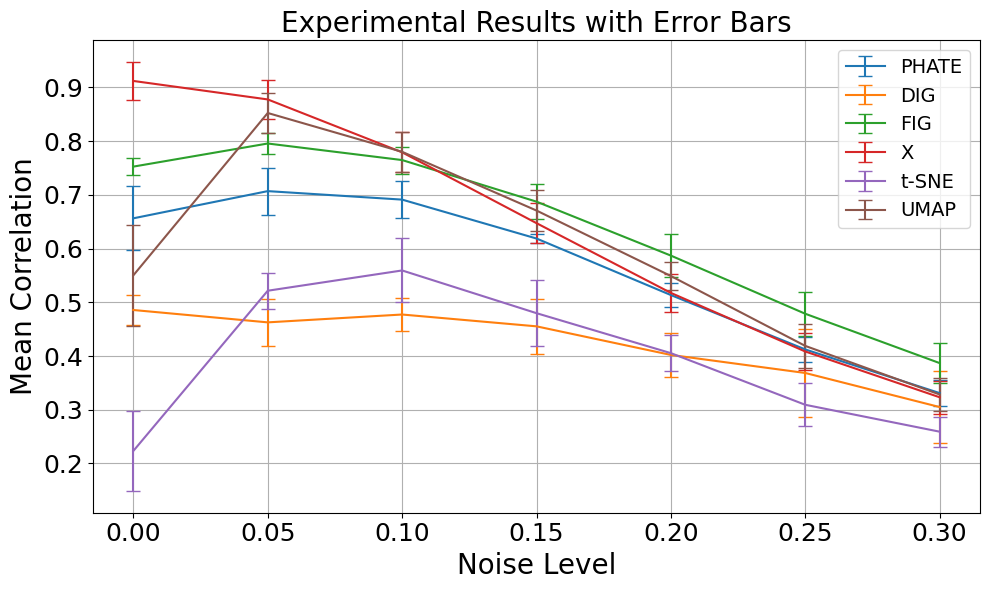}
  \caption{Mantel coefficient between different embedding distances and the ground truth parameters $\boldsymbol{\theta}$ of the simulated random walk. FIG outperforms all methods in the high noise setting and is competitive in the low noise setting.}
  \label{fig:mantelofrandomwalk}
\end{figure}

To evaluate global disparities among embeddings, we employed the Mantel~\cite{mantel1967detection} test. The Mantel test produces correlation coefficients ranging from $0$ to $1$ between two sets of distances, similar to the standard Pearson correlation. However, the Mantel test takes into account the interdependence of distances, recognizing that altering the position of a single observation impacts $N - 1$ distances, challenging the adequacy of a simple correlation calculation for evaluating similarity between distance matrices.  To compute the Mantel correlation coefficients, we initially compute the pairwise Euclidean distances of the embeddings and the pairwise Euclidean distances of the dynamical process $\boldsymbol{\theta}$, then calculate the Mantel correlations between these two distance matrices. To ensure reproducibility, for each noise level, we tested using 5 different random seeds to generate the data. We present the mean Mantel correlations (as solid lines) along with the standard deviations (as error bars) in Figure~\ref{fig:mantelofrandomwalk}. We also include another line (blue) in which the noisy data  is used to compute the Mantel correlation with $\boldsymbol{\theta}$.

From Figure~\ref{fig:mantelofrandomwalk}, we observe that as the noise level increases, the Mantel coefficient between the data and $\boldsymbol{\theta}$ drops significantly. Initially, in the noiseless setting, none of the methods outperform the original data in terms of the Mantel coefficient, although FIG produces the best result out of all the embedding methods. As the noise level rises, all methods tend to have lower Mantel coefficients with $\boldsymbol{\theta}$, but FIG outperforms the others at higher noise levels. When the noise level reaches 0.15, the Mantel coefficient of FIG is even higher than that of the original data, indicating that FIG is more noise resilient compared to other approaches. In particular, the performance gap between FIG and DIG is fairly large. This may be because DIG estimates histograms in high dimensions which  may not be accurate when the data is complex and noisy, leading to over-smoothing in the simulated steps. In contrast, FIG uses basis functions for the new features, avoiding density estimation in higher dimensions and thus preventing over-smoothing in the random walk.

\subsection{Visualizing Sleep Patterns -- EEG}
\begin{figure*}[h!]
\centering
\begin{subfigure}{1\textwidth}
  \centering
  \caption*{\textbf{(a)} Embeddings \label{fig:eeg_subfig_embeddings}} 
  \includegraphics[width=1.0\linewidth]{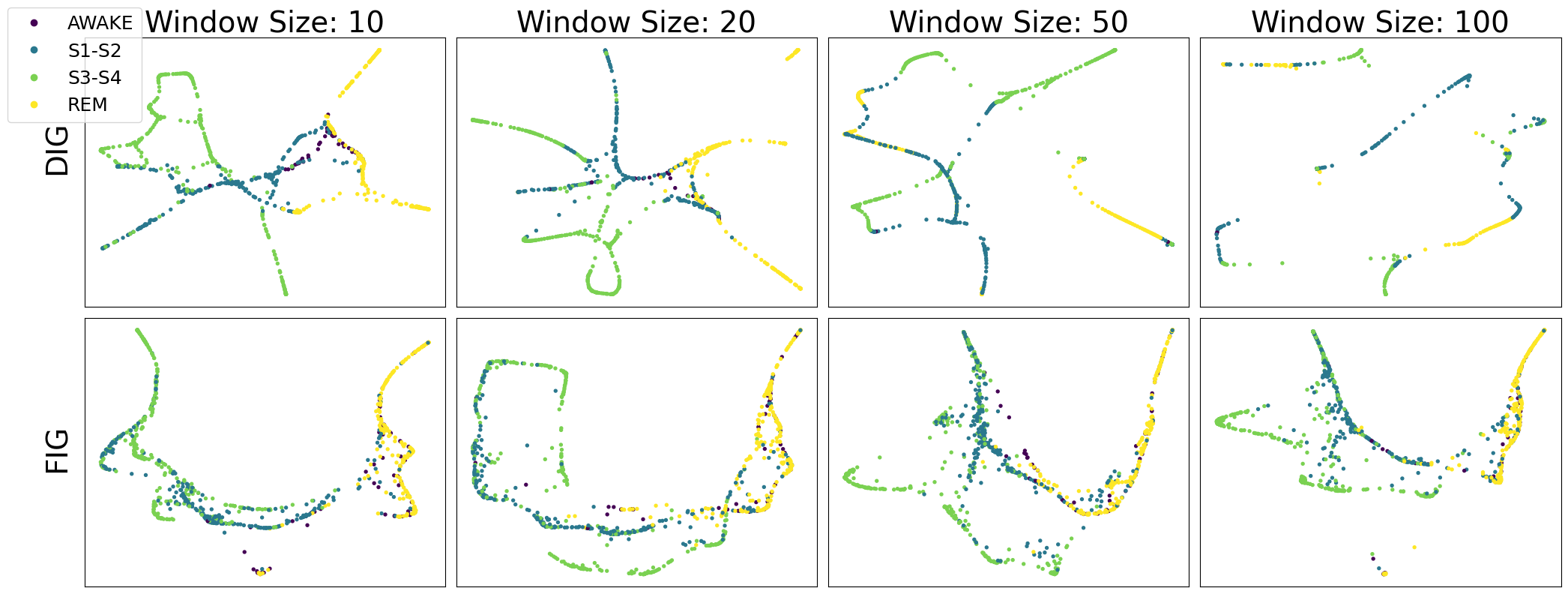}
\end{subfigure}
\begin{subfigure}{0.5\textwidth}
  \centering
  \caption*{\textbf{(b)} Mantel of DIG}
  \label{fig:eeg_subfig_mantel_dig}
  \includegraphics[width=0.8\linewidth]{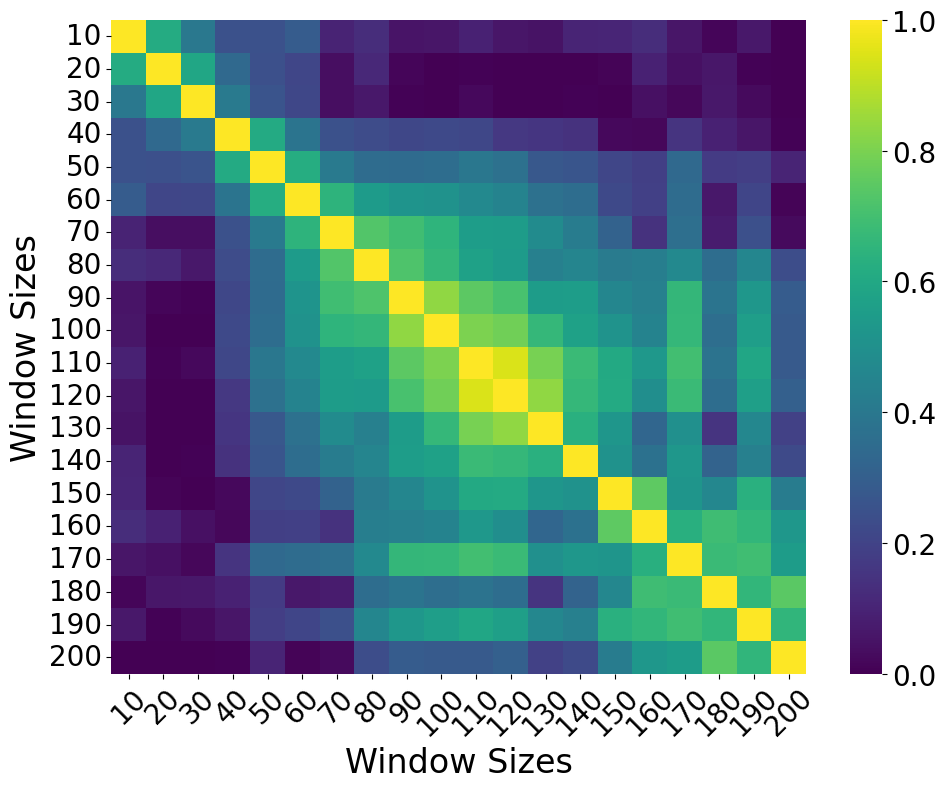}
\end{subfigure}%
\hspace*{\fill} 
\begin{subfigure}{0.5\textwidth}
  \centering
  \caption*{\textbf{(c)} Mantel of FIG}
  \label{fig:eeg_subfig_mantel_fig}
  \includegraphics[width=0.8\linewidth]{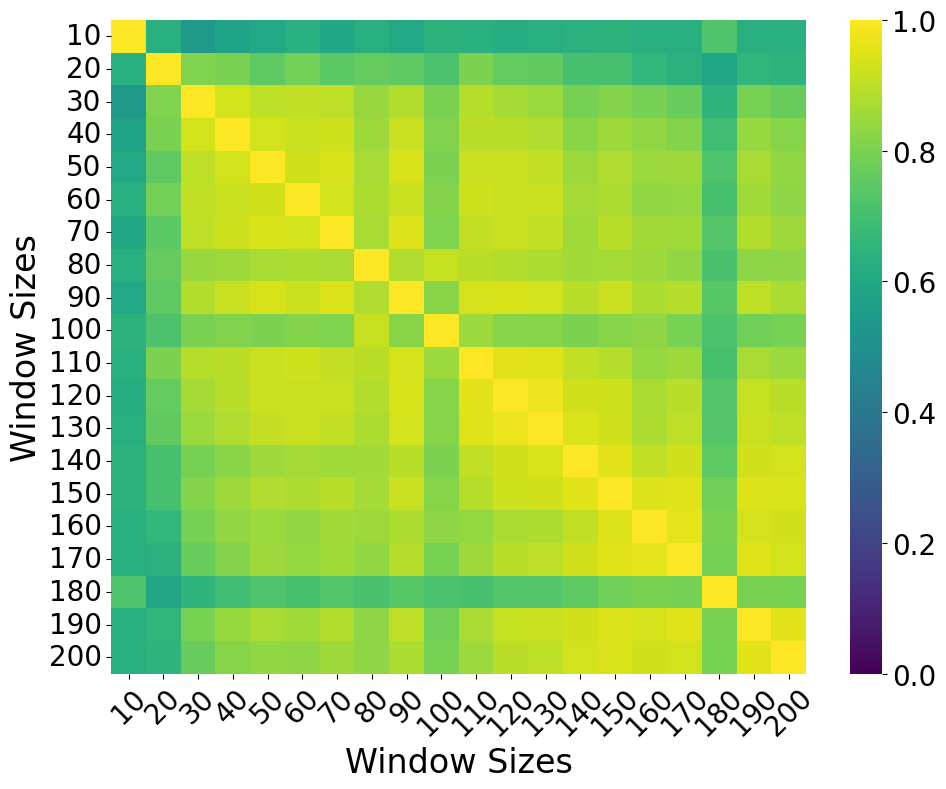}
\end{subfigure}
\caption{Comparison of FIG and DIG on EEG brain measurements during different sleep stages . FIG is more robust to different window sizes than DIG. (a) A visual comparison of FIG and DIG. Parts (b) and (c) show pairwise Mantel correlations between the embeddings.}
\label{fig:eegresults}
\end{figure*}
We now apply FIG to EEG data sourced from~\cite{eeg, eeg1}. The initial data is a multivariate time series data with 18 dimensions, sampled at 512Hz, is classified into six sleep stages following R\&K rules (REM, Awake, S-1, S-2, S-3, S-4), each covering a 30-second period. To address limited observations within certain stages, we combine S-1 with S-2 and S-3 with S-4. Furthermore, the data undergoes band filtering within the 8-40 Hz range and is subsequently down-sampled to 128Hz. Given the substantial size of the original data (exceeding 3 million samples), we preprocess it by computing Fourier basis functions and subsequently averaging the functions $\phi(\x_t)$ over a segment.Following Algorithm~\ref{FigAlgorithm}, we use seven Fourier basis functions for each dimension, adopting identical settings as in DIG~\cite{dig}, we set $\mathcal{L}_1 = 3840$ and the distance between histogram centers is also $\mathcal{L}_1$, corresponding to the number of observations within a 30-second interval. 
Finally, we obtain the embeddings in a 2-dimensional space using Algorithm~\ref{FigAlgorithm}.

As a baseline, we  applied PHATE, Diffusion Maps, UMAP, t-SNE to the data. None of these methods were able to capture any meaningful structure (see Figure~\ref{fig:eeg_subfig_embeddings_others} in the Appendix). Thus we focus our comparisons on FIG and DIG. Using the same settings from~\cite{dig}, we selected the number of bins for the histogram estimation $Nb = 20$ per dimension for DIG, and varied window sizes $\mathcal{L}_2 = 10, 20, .... 200$ for both FIG and DIG to compute the embeddings. In Figure~\ref{fig:eegresults} part(a), we illustrate the 2-D embeddings of FIG alongside DIG (the baseline) for various window sizes $\mathcal{L}_2$. Figure~\ref{fig:eegresults} part (b) and part (c) present the mean pairwise Mantel correlations of DIG and FIG respectively across 20 window sizes, ranging from 10 to 200, comparing the 2-D embeddings of DIG and FIG. 5 different seeds were used.  The relatively small standard deviations of the Mantel correlations indicate the reproducibility of our experiments (see  Figure~\ref{fig:eeg_std} in the Appendix).

From Figure~\ref{fig:eegresults}(a), we observe qualitatively that as the window size increases, the 2-dimensional embeddings of DIG tend to lose important structural information such as the connections between different sleep stages. In contrast, the embeddings of FIG remain stable while providing similar branching and trajectory structures as DIG with smaller window sizes. This suggests that FIG is more resilient to the choice of window size than DIG. Similar results are observed in the 3D embeddings (Figure~\ref{fig:eeg_subfig_embeddings_3d} in the Appendix).
Figures~\ref{fig:eegresults}(b) and (c) corroborate this numerically. Here we observe that the FIG embeddings with different window sizes have higher Mantel correlation coefficients with each other than DIG. On the other hand, we observe a high Mantel correlation for FIG among different window sizes, which corresponds to the visualization results observed in part (a). We also demonstrated that the computational time of FIG is optimal compared to DIG, as shown in Table~\ref{table:computationtime} in the Appendix.

\section{Conclusion}
\label{sec:conclusion}
Our contribution lies in the development of a novel visualization method, FIG, based on the functional data framework and integrating the dynamic framework of EIG-based metrics~\cite{eig,eig1}. Unlike EIG, our approach bypasses density estimation, which is particularly beneficial in high-dimensional data settings. Extensive experiments demonstrate the effectiveness of FIG in achieving robust visualizations, with stability observed across different window lengths for computing Mahalanobis distances. One limitation of the experiments in this paper is their application solely to time series data. This limitation arises from the necessity to define a "time window," which poses challenges in non-time series contexts. However, we established in Section~\ref{sec:FIG} that FIG can be extended to non-time series data via a proper definition of neighbors between points. We leave experimental validation of this for future work.

\bibliographystyle{IEEEtran}
\bibliography{references}

\newpage
\appendix
\section*{Appendix}
\label{sec:appendix}
Here we provide additional figures from our experiments on the EEG data as well as a computational comparison between DIG and FIG in Table~\ref{table:computationtime}.

\begin{figure}[h!]
  \centering
  \includegraphics[width=1.0\linewidth]{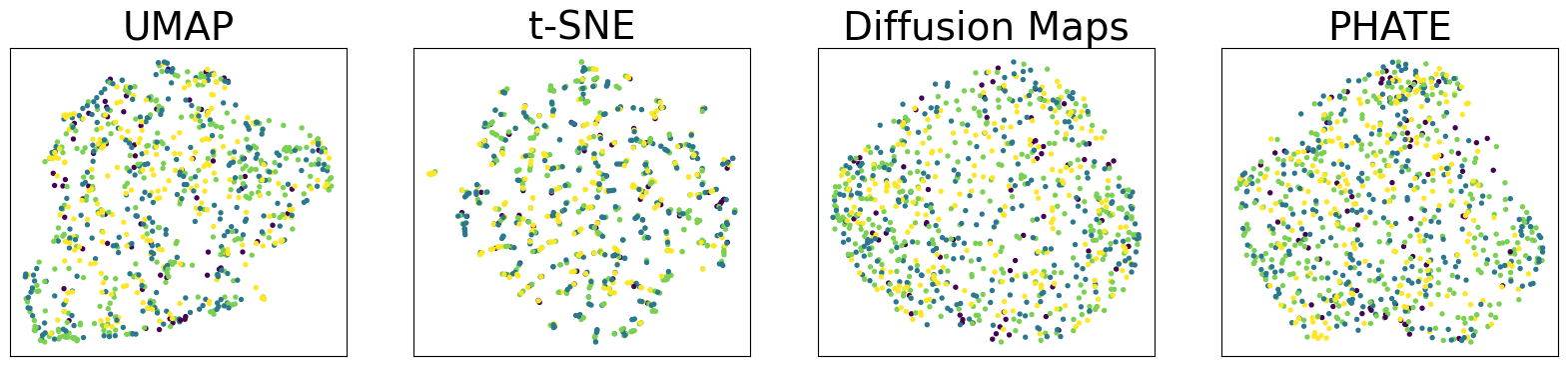}
  \caption{2D embeddings of the EEG data using other methods, colored by the same labels (sleep stages). These methods are unable to capture any of the structure in the data, in contrast with both FIG and DIG (Figure~\ref{fig:eegresults}).}
  \label{fig:eeg_subfig_embeddings_others}
\end{figure}

\begin{figure*}[h!]
\begin{subfigure}{0.5\textwidth}
  \centering
  \includegraphics[width=0.9\linewidth]{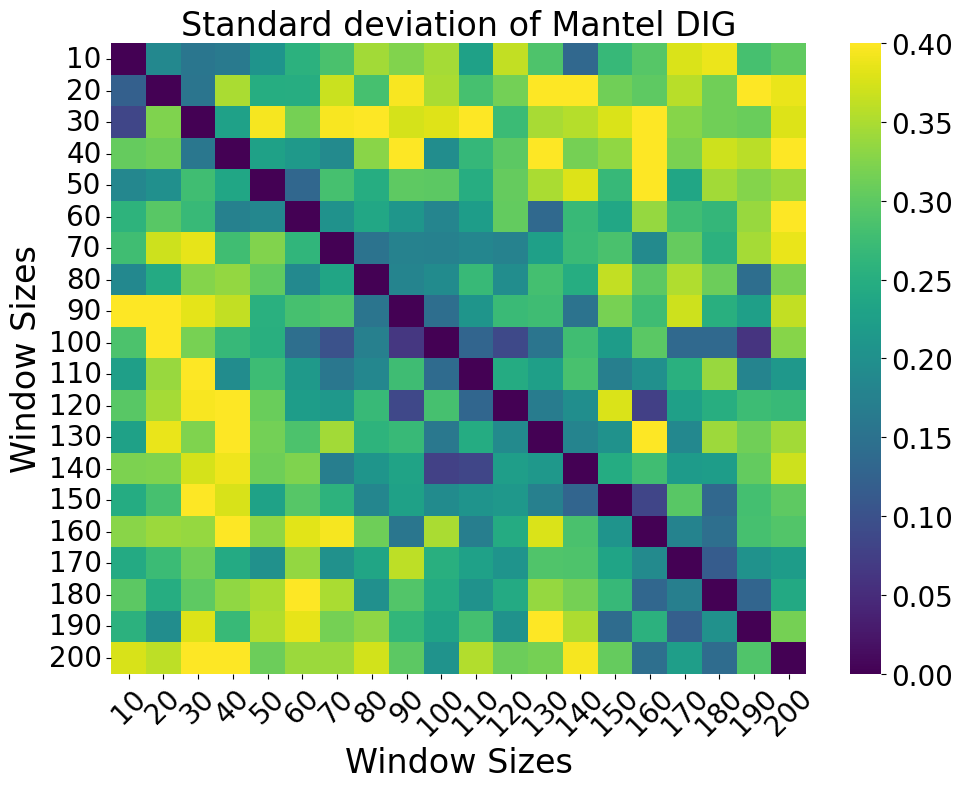}
\end{subfigure}%
\hspace*{\fill} 
\begin{subfigure}{0.5\textwidth}
  \centering
  \includegraphics[width=0.9\linewidth]{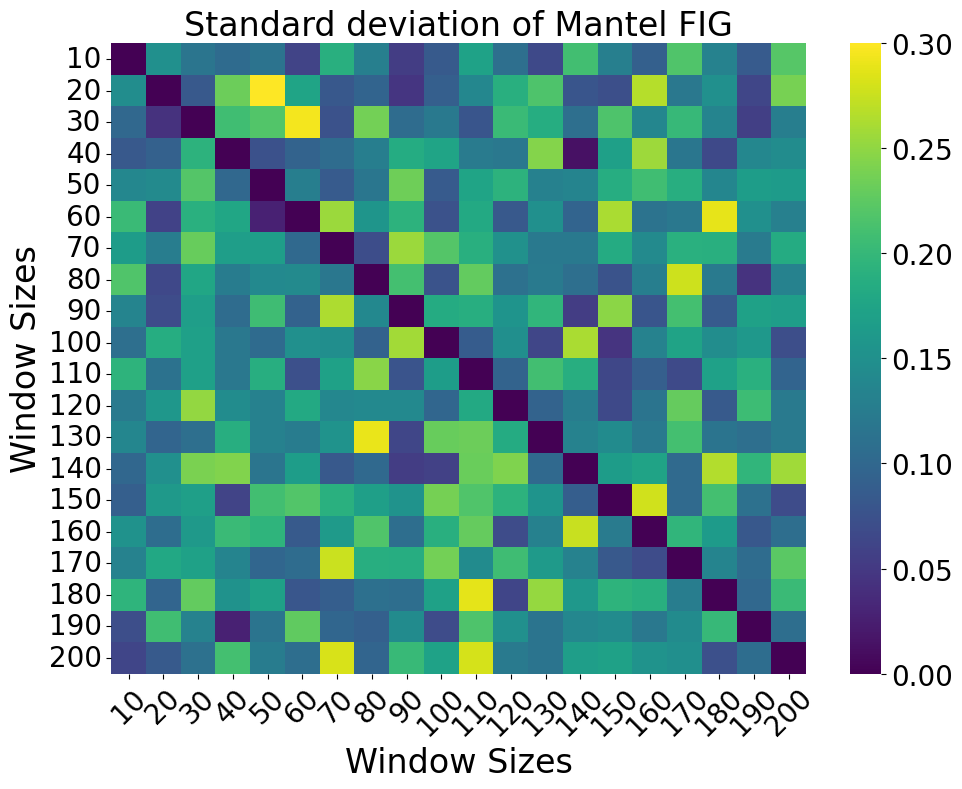}
\end{subfigure}
\caption{The standard deviations of Mantel test results for DIG and FIG on the EEG data computed from  5 runs. The corresponding average Mantel correlations are given in Figure~\ref{fig:eegresults}.  The standard deviations are generally low, especially for FIG, indicating our results are reproducible.}
\label{fig:eeg_std}
\end{figure*}

\begin{figure}[h!]
  \centering
  \includegraphics[width=1.0\textwidth]{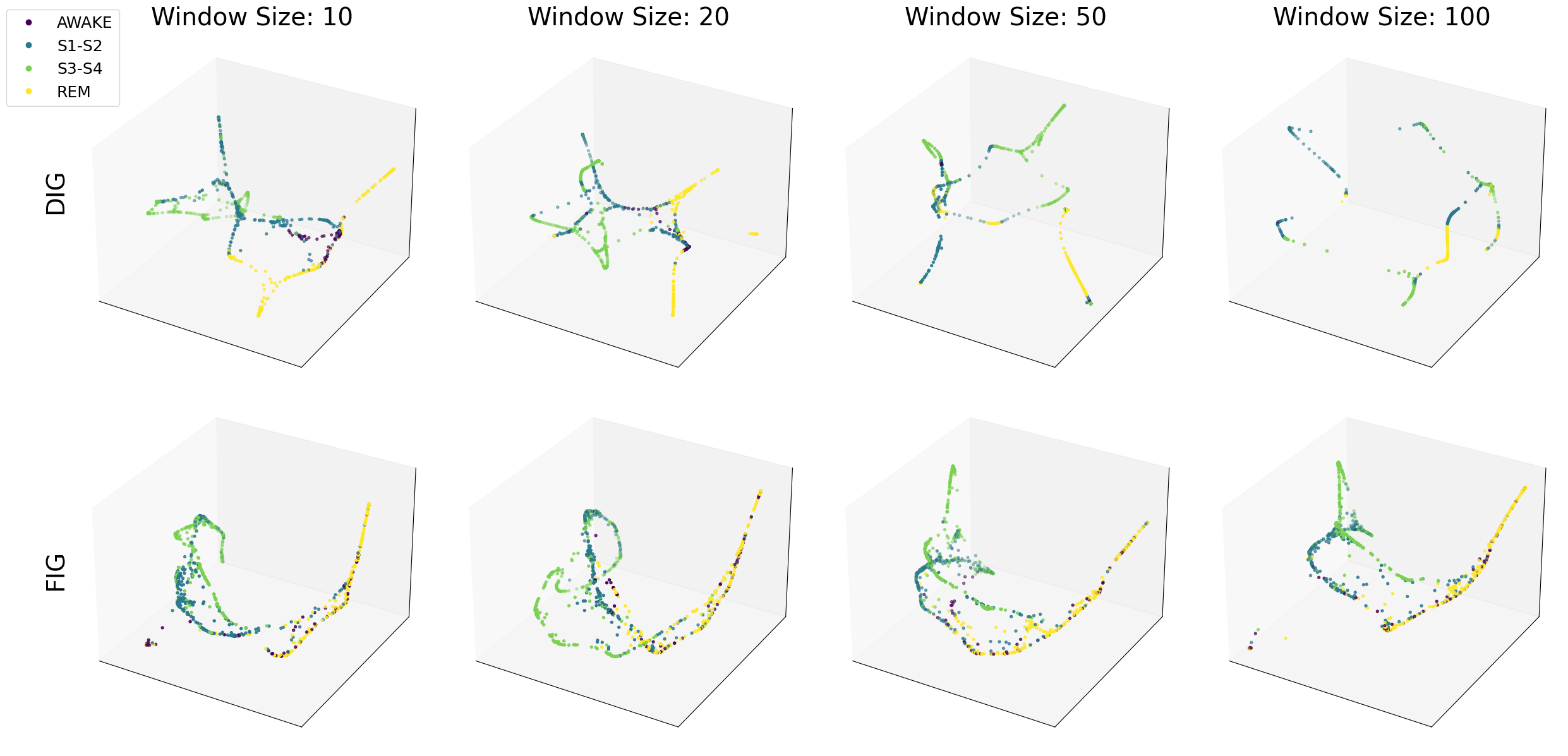}
  \caption{3D embeddings of the EEG data comparing DIG versus FIG for various window sizes. The FIG embeddings are generally more robust to the window size than DIG.} 
  \label{fig:eeg_subfig_embeddings_3d}
\end{figure}

\begin{table}[]
\caption{Computation time of the Mahalanobis distances for DIG and FIG in seconds over 5 different seeds of EEG data for window size $\mathcal{L}_2 = 10$.We anticipate minimal variance of the time across varying window sizes, given that $\mathcal{L}_2$ is employed in computing the covariance matrix after the density estimation. FIG is much faster than DIG.}
\label{table:computationtime}
\centering
\begin{tabular}{||c c c ||} 
 \hline
 Seeds & DIG & FIG \\ [0.5ex] 
 \hline\hline
 1 & 171.2 & 33.2\\ 
 \hline
 2 & 174.4 & 35.2 \\
 \hline
 3 & 161.3 & 23.2 \\
 \hline
 4 & 157.5 & 25.7 \\
 \hline
 5 & 161.7 & 34.2\\
 \hline
\end{tabular}
\end{table}

\end{document}